\def\year{2020}\relax
\documentclass[letterpaper]{article} 
\usepackage{aaai20}  
\usepackage{times}  
\usepackage{helvet} 
\usepackage{courier}  
\usepackage[hyphens]{url}  
\usepackage{graphicx} 
\usepackage{graphicx}  
\usepackage{booktabs}
\usepackage{tabularx}
\usepackage{array}
\usepackage{amsmath}
\usepackage{amsfonts}
\usepackage[dvipsnames]{xcolor}
\usepackage{subfigure}
\usepackage{colortbl}
\newcolumntype{C}[1]{>{\centering\let\newline\\\arraybackslash}p{#1}}
\newcolumntype{M}[1]{>{\centering\let\newline\\\arraybackslash}m{#1}}

\title{An Adversarial Perturbation Oriented Domain Adaptation Approach for Semantic Segmentation}
\author{Jihan Yang$^{1,2}$, Ruijia Xu$^{1}$, Ruiyu Li$^{2}$, Xiaojuan Qi$^{3}$, Xiaoyong Shen$^{2}$, Guanbin Li$^{1}$\thanks{Corresponding author is Guanbin Li. This work was supported in
part by the State Key Development Program under Grant No.2016YFB1001004, in part by the National Natural Science Foundation of China under Grant No.61976250 and No.U1811463, in part by the Fundamental Research Funds for the Central Universities under Grant No.18lgpy63.  This work was also supported by SenseTime Research Fund.}, Liang Lin$^{1,4}$\\
$^{1}$School of Data and Computer Science, Sun Yat-sen University, China \\
$^{2}$Tencent YouTu Lab, $^{3}$University of Oxford\\ 
$^{4}$DarkMatter AI Research \\
\{jihanyang13, ruijiaxu.cs\}@gmail.com, \{royryli, dylanshen\}@tencent.com xiaojuan.qi@eng.ox.ac.uk\\
liguanbin@mail.sysu.edu.cn, linliang@ieee.org}

\begin{document}

\maketitle

\begin{abstract}
    We focus on Unsupervised Domain Adaptation (\textit{UDA}) for the task of semantic segmentation. 
    Recently, adversarial alignment has been widely adopted to match the marginal distribution of feature representations across two domains \textit{globally}.
    However, this strategy fails in adapting the representations of the tail classes or small objects for semantic segmentation since the alignment objective is dominated by head categories or large objects.
    In contrast to adversarial alignment, we propose to explicitly train a domain-invariant classifier by generating and defensing against \textit{pointwise} feature space adversarial perturbations.
    Specifically, we firstly perturb the intermediate feature maps with several attack objectives (i.e., discriminator and classifier) on each individual position for both domains, and then the classifier is trained to be invariant to the perturbations.
    By perturbing each position individually, our model treats each location evenly regardless of the category or object size and thus circumvents the aforementioned issue. Moreover, the domain gap in feature space is reduced by extrapolating source and target perturbed features towards each other with attack on the domain discriminator.
    Our approach achieves the state-of-the-art performance on two challenging domain adaptation tasks for semantic segmentation: GTA5 $\rightarrow$ Cityscapes and SYNTHIA $\rightarrow$ Cityscapes.
\end{abstract}

\section{Introduction}
\begin{figure}[t]
    \centering
    \setlength{\abovecaptionskip}{0.1cm}
    \setlength{\belowcaptionskip}{-0.4cm}
    \subfigure[RGB image]{
        \begin{minipage}[t]{0.474\linewidth}
            \includegraphics[width=\textwidth]{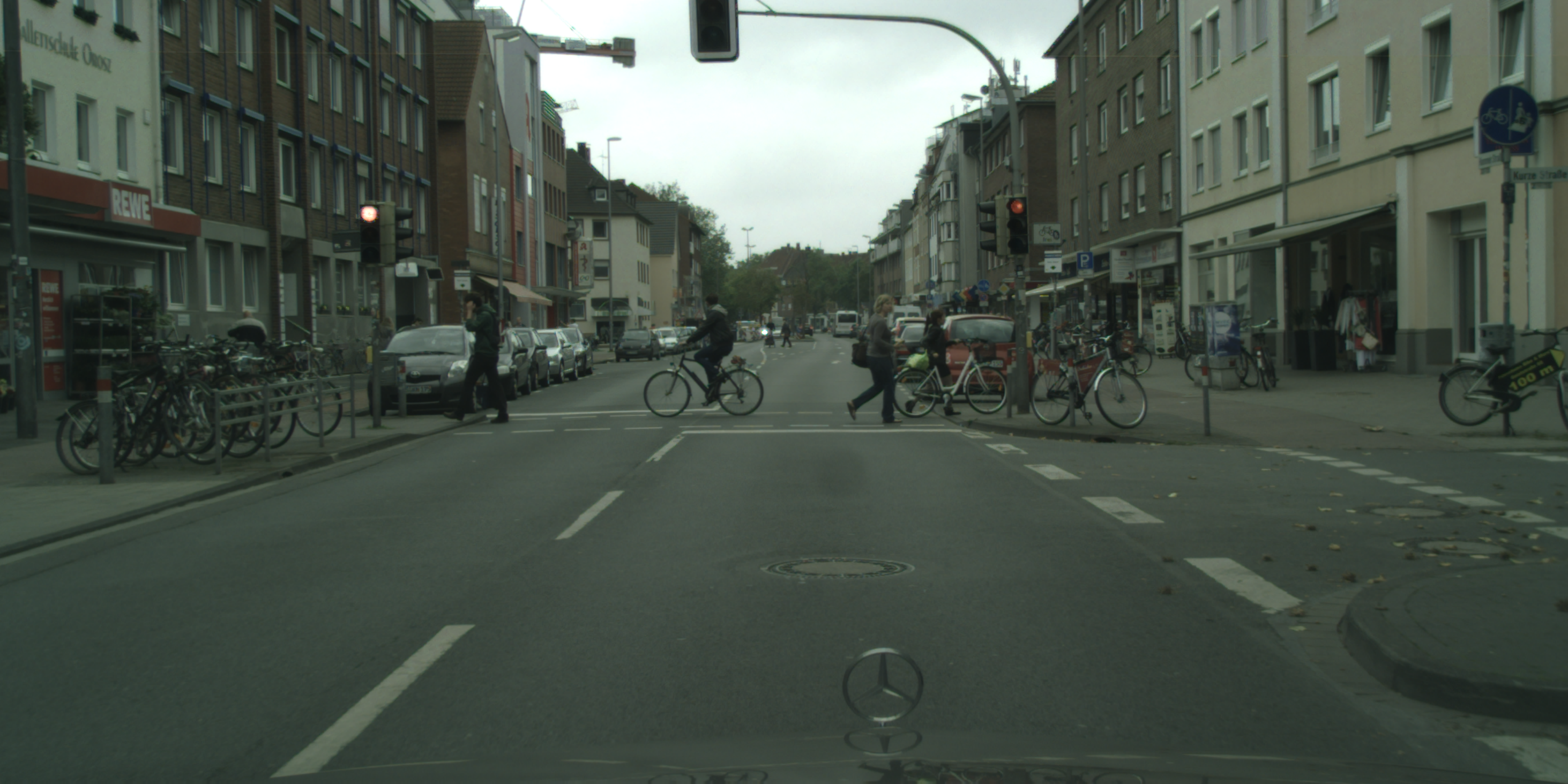}
            \label{fig:1-a}
        \end{minipage}
    }
    \subfigure[Without adaptation]{
        \begin{minipage}[t]{0.474\linewidth}
            \includegraphics[width=\textwidth]{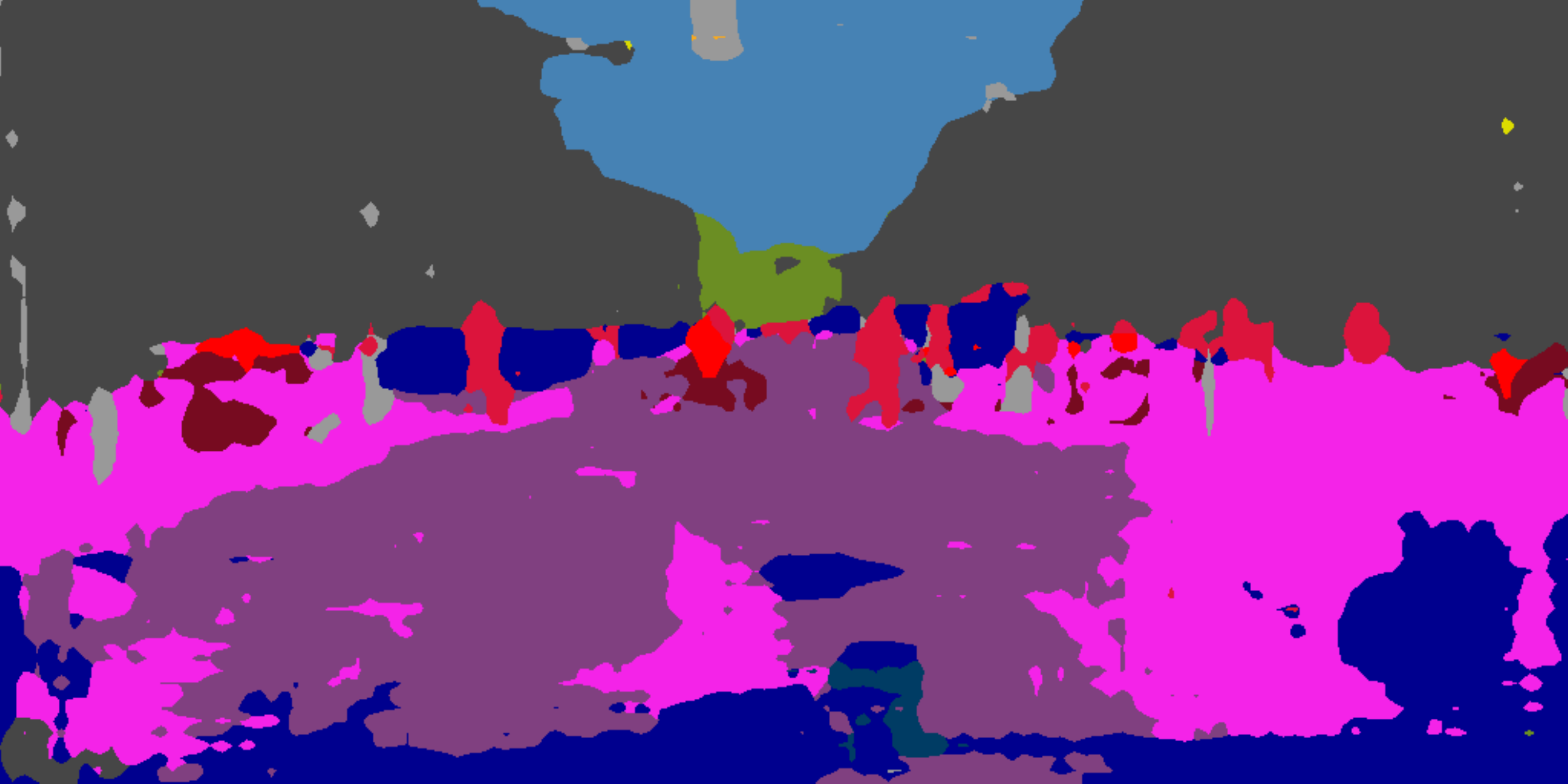}
            \label{fig:1-b}
        \end{minipage}
    }
    \subfigure[ASN (Tsai et al. 2018)]{
        \begin{minipage}[t]{0.474\linewidth}
            \includegraphics[width=\textwidth]{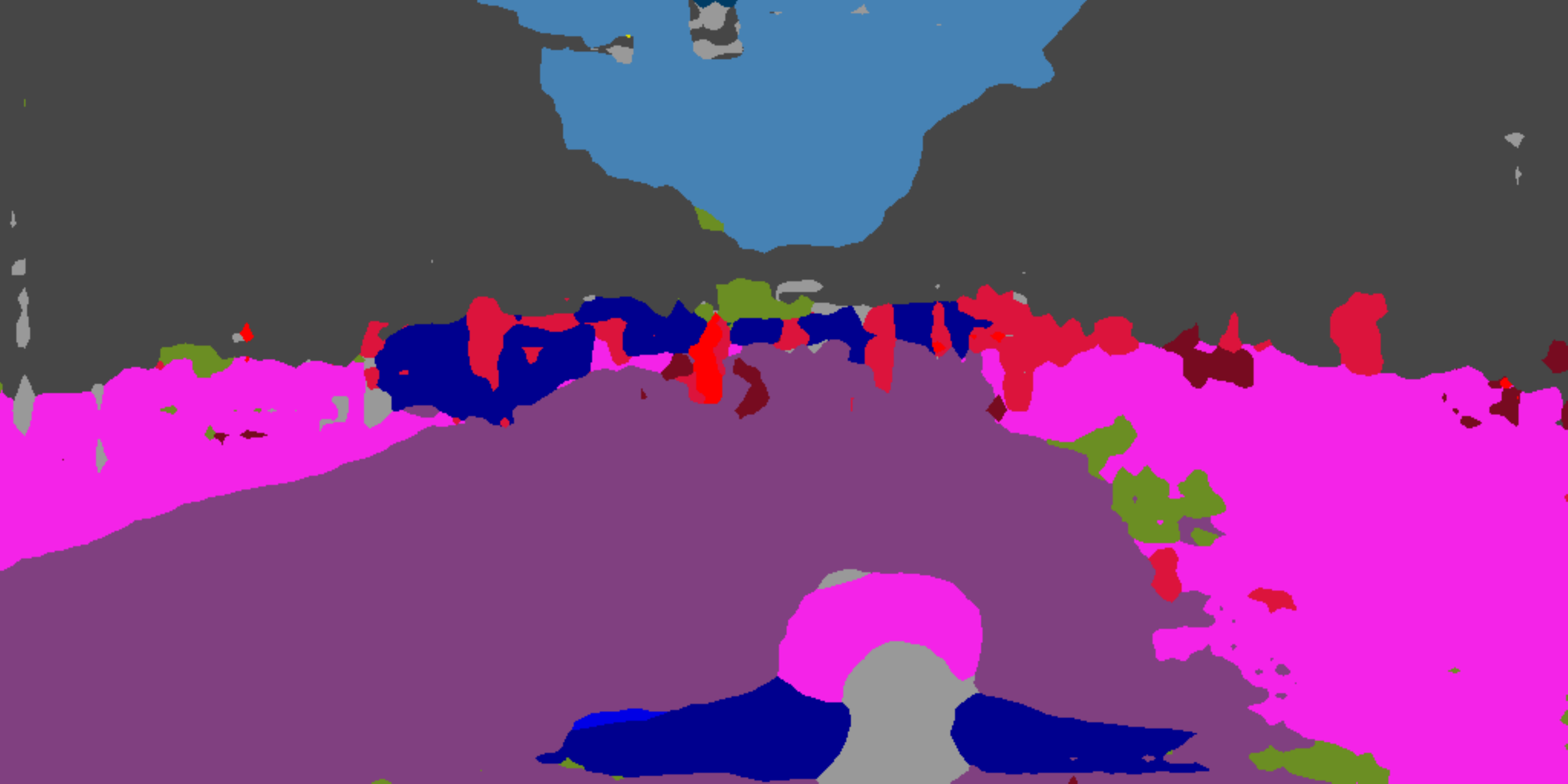}
            \label{fig:1-c}
        \end{minipage}
    }
    \subfigure[Ours]{
        \begin{minipage}[t]{0.474\linewidth}
            \includegraphics[width=\textwidth]{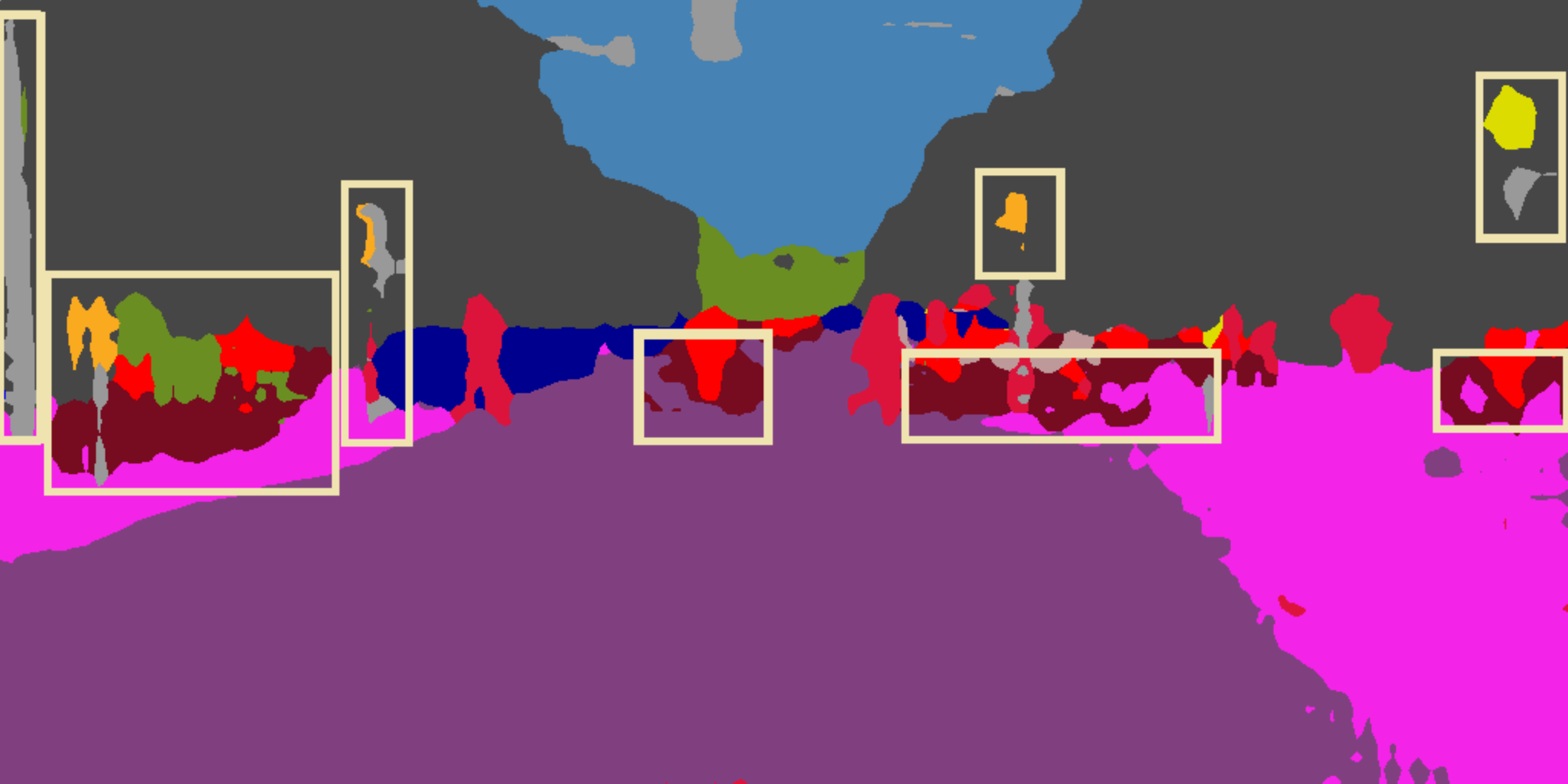}
            \label{fig:1-d}
        \end{minipage}
    }
    \caption{Comparison of semantic segmentation output. This example shows our method can evenly capture information of different categories, while classical adversarial alignment method such as ASN~\cite{tsai2018learning} might collapse into {head (i.e., background) classes or large objects}.}
    \label{fig:result_comparison}
\end{figure}

Semantic segmentation is a fundamental problem in computer vision with many applications in robotics, autonomous driving, medical diagnosis, image editing, etc. The goal is to assign each pixel with a semantic category. Recently, this field has gained remarkable progress via training deep convolutional neural networks (CNNs) \cite{long2015fully} on large scale human annotated datasets \cite{Cordts_2016_CVPR}. 
However, models trained on specific datasets may not generalize well to novel scenes (see Figure~\ref{fig:result_comparison}(b)) due to the inevitable visual domain gap between training and testing datasets. This seriously limits the applicability of the model in diversified real-world scenarios.
For instance, an autonomous vehicle might not be able to sense its surroundings in a new city or a changing weather condition. 
To this end, learning domain-invariant representations for semantic segmentation has drawn increasing attentions.   
\par

Towards the above goal, Unsupervised Domain Adaptation (\textbf{UDA}) has shown promising results \cite{vu2019advent,luo2019taking}.
UDA aims to close the gap between the annotated source domain and unlabeled target domain by learning domain-invariant while task-discriminative representations.
Recently, adversarial alignment has been recognized as an effective way to obtain such representations~\cite{hoffman2016fcns,hoffman2017cycada}.
Typically, in adversarial alignment, a discriminator is trained to distinguish features or images from different domains, while the deep learner tries to generate features to confuse the discriminator. Recent representative approach ASN~\cite{tsai2018learning} is proposed to match the source and target domains in the output space and has achieved promising results.\par

However, adversarial alignment based approaches can be easily overwhelmed by dominant categories (i.e., background classes or large objects).
Since the discriminator is only trained to distinguish the two domains globally, it can not produce category-level or object-level supervisory signal for adaptation. Thus, the generator is not enforced to evenly capture category-specific or object-specific information and fails to adapt representations for the tail categories.
We term this phenomenon as \textit{category-conditional shift} and highlight it in the Figure~\ref{fig:result_comparison}. ASN performs well in adapting head categories (e.g., road) and gains improvement when viewed globally, but fails to segment the tail categories such as ``sign'', ``bike''.
Missing the small instances (e.g., traffic light) is generally intolerable in real-world applications.
While we can moderate this issue by equipping the segmentation objective with some heuristic re-weighting schemes \cite{berman2018lovasz}, those solutions usually rely on implicit assumptions about the model or the data (e.g., L-Lipschitz condition, overlapping support \cite{wu2019domain}), which are not necessarily met in real-world scenarios. In our case, we empirically show that the adaptability achieved by those approximate strategies are sub-optimal.

In this paper, we propose to perform domain adaptation via feature space adversarial perturbation inspired by~\cite{goodfellow2014explaining}. Our approach mitigates the category-conditional shift by iteratively generating pointwise adversarial perturbations and then defensing against them for both the source and target domains. 
Specifically, we firstly perturb the feature representations for both source and target samples by appending gradient perturbations to their original features. The perturbations are derived with adversarial attacks on the discriminator to assist in filling in the representation gap between source and target, as well as the classifier to capture the vulnerability of the model. This procedure is facilitated with the proposed Iterative Fast Gradient Sign Preposed Method (I-FGSPM) to mitigate the huge gradient gap among multiple attack objectives.
Taking the original and perturbed features as inputs, the classifier is further trained to be domain-invariant by defensing against the adversarial perturbations, which is guided by the source domain segmentation supervision and the target domain consistency constraint.

Instead of aligning representations across domains globally, our perturbation based strategy is conducted on each individual position of the feature maps, and thus can capture the information of different categories evenly and alleviate the aforementioned category-conditional shift issue. 
In addition, the adversarial features also capture the vulnerability of the classifier, thus the adaptability and capability of the model in handling hard examples (typically tail classes or small objects) is further improved by defensing against the perturbations.
Furthermore, since we extrapolate the source adversarial features towards the target representations to fill in the domain gap, our classifier can be aware of the target features as well as receiving source segmentation supervision, which further promotes our classifier to be domain-invariant.
Extensive experiments on GTA5 $\rightarrow$ Cityscapes and SYNTHIA $\rightarrow$ Cityscapes have verified the state-of-the-art performance of our method.

\begin{figure*}[h]
    \centering
    \includegraphics[width=\textwidth]{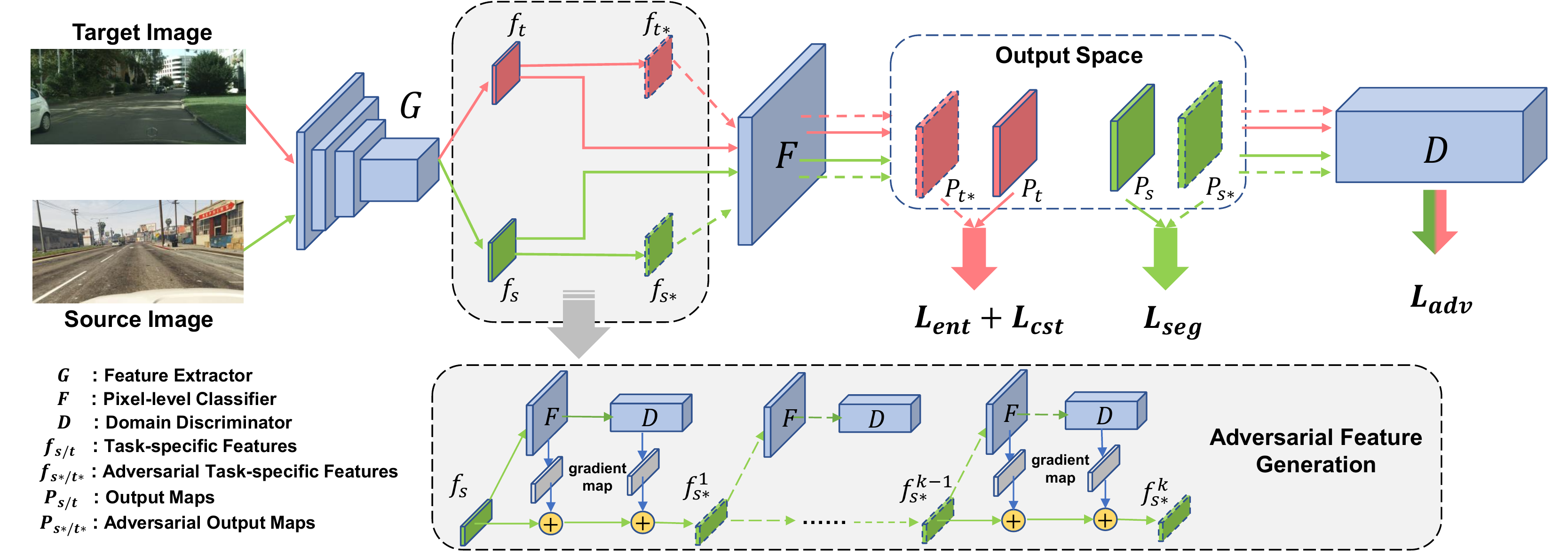}
    \caption{Framework Overview. We illustrate step \textcolor{red}{2} in the shaded area where source features are taken as an example. In light of $f_{s/t}$ extracted from the feature extractor $G$, we employ the multi-objective adversarial attack with our proposed I-FGSPM on the classifier $F$ as well as discriminator $D$ and then accumulate the gradient maps. Therefore, we obtain the mutated features $f_{s*/t*}$ after appending the perturbations to the original copies. Furthermore, these perturbed and original features are trained by an adversarial training procedure (i.e., step \textcolor{red}{3}), which is presented in the upper right. We have highlighted the different training objectives for the output maps of their corresponding domains, which are predicted by the classifier $F$ and then followed by the discriminator $D$ to produce domain prediction maps. The \textcolor{green}{green} and \textcolor{OrangeRed}{red} colors stand for the source and target flows respectively.}
    \label{fig:framework}
\end{figure*}

\section{Related Work}
\textbf{Semantic Segmentation} is a highly active and important research area in visual tasks. Recent fully convolutional network based methods~\cite{chen2017deeplab,zhao2017pyramid} have achieved remarkable progress in this field by training deep convolutional neural networks on numerous pixel-wise annotated images. However, building such large-scale datasets with dense annotations takes expensive human labor. An alternative approach {is to train} model on synthetic data (e.g., GTA5~\cite{richter2016playing}, SYNTHIA~\cite{ros2016synthia}) and transfer to real-world data. Unfortunately, even a subtle departure from the training regime can cause catastrophic model degradation when generalized into new environments. The reason lies in the different data distributions between source and target domains, known as domain shift.\\
\textbf{Unsupervised Domain Adaptation} approaches have achieved remarkable success in addressing aforementioned problem. Existing methods mainly focus on minimizing the statistic distance such as Maximum Mean Discrepancy~(MMD) of two domains~\cite{long2015learning,long2017deep}. Recently, inspired by GAN~\cite{goodfellow2014generative}, adversarial learning is successfully explored to entangle feature distributions from different domains~\cite{ganin2014unsupervised,ganin2016domain}. \citeauthor{hoffman2016fcns}~\shortcite{hoffman2016fcns} applied feature-level adversarial alignment method in UDA for semantic segmentation. Several following works improved this framework for pixel-wise domain adaption~\cite{chen2017no,chen2018road}. Besides alignment in the bottom feature layers, \citeauthor{tsai2018learning}~\shortcite{tsai2018learning} found that output space adaptation via adversarial alignment might be more effective. \citeauthor{vu2019advent}~\shortcite{vu2019advent} further proposed to align output space entropy maps. On par with feature-level and output space alignment methods, the remarkable progress of unpaired image to image translation~\cite{zhu2017unpaired} inspired several methods to address pixel-level adaptation problems~\cite{hoffman2017cycada,zhang2018fully}. Among some other approaches, \citeauthor{zou2018unsupervised}~\shortcite{zou2018unsupervised} used self-training strategy to generate pseudo labels for unlabeled target domain. \citeauthor{saito2017asymmetric}~\shortcite{saito2017asymmetric} utilized tri-training to assign pseudo labels and obtain target-discriminative representations, while \citeauthor{luo2019taking}~\shortcite{luo2019taking} proposed to compose tri-training and adversarial alignment strategies to enforce category-level feature alignment. And \citeauthor{saito2018maximum}~\shortcite{saito2018maximum} used two-branch classifiers and generator to minimize $H\Delta H$ distance. Recently,~\citeauthor{Xu_2019_ICCV}~\shortcite{Xu_2019_ICCV} reveals that progressively adapting the task-specific feature norms of the source and target domains to a large range of values can result in significant transfer gains.\\
\textbf{Adversarial Training} injects perturbed examples into training data to increase robustness. These perturbed examples are designed for fooling machine learning models. To the best of our knowledge, adversarial training strategy is originated in \cite{szegedy2013intriguing} and further studied by \citeauthor{goodfellow2014explaining}~\shortcite{goodfellow2014explaining}. Several attack methods are further designed for efficiently generating adversarial examples \cite{kurakin2016adversarial,dong2018boosting}. As for UDA, \citeauthor{volpi2018generalizing}~\shortcite{volpi2018generalizing} generated adversarial examples to adaptively augment the dataset. \citeauthor{liu2019transferable}~\shortcite{liu2019transferable} produced transferable examples to fill in the domain gap and adapt classification decision boundary.
However, the above approach is only validated on the classification task for unsupervised domain adaptation. Our approach shares similar spirit with \citeauthor{liu2019transferable}, while we investigate adversarial training in the field of semantic segmentation to generate pointwise perturbations that improve the robustness and domain invariance of the learners.

\section{Method}
Considering the problem of unsupervised domain adaptation in semantic segmentation.
Formally, we are given a source domain $ \mathcal{S}$ and a target domain $\mathcal{T}$. We have access to the source data $x_s \in \mathcal{S}$ with pixel-level labels $y_s$ and the target data $x_t \in \mathcal{T}$ without labels.
Our overall framework is shown in Figure~\ref{fig:framework}. Feature extractor $G$ takes images $x_s$ and $x_t$ as inputs and produces intermediate feature maps $f_s$ and $f_t$; Classifier $F$ takes features $f_s$ and $f_t$ from $G$ as inputs and predicts $C$-dimensional segmentation softmax outputs $P_s$ and $P_t$; Discriminator $D$ is a CNN-based binary classifier with a fully-convolutional output to distinguish whether the input ($P_s$ or $P_t$) is from the source or target domain.

To address aforementioned category-conditional shift, we propose a framework that alternatively generates pointwise perturbations with multiple attack objectives and defenses against these perturbed copies via an adversarial training procedure. Since our framework conducts perturbations for each point independently, it circumvents the interference of different categories. Our learning procedure can also be seen as a form of active learning or hard example mining, where the model is enforced to minimize the worst case error when features are perturbed by adversaries. Our framework consists of three steps as follow: \par
\textbf{Step 1: Initialize $G$ and $F$.} We train both the feature extractor $G$ and classifier $F$ with source samples. Since we need $G$ and $F$ to learn task-specific feature representations, this step is crucial.
Specifically, we train the feature extractor and classifier by minimizing cross entropy loss as follow:
\begin{equation}
    L_{ce}(x_s, y_s) = - \sum_{h=1}^{H}\sum_{w=1}^{W} \sum_{c=1}^{C} y_s^{(h,w,c)} \log P_{s}^{(h,w,c)},
\end{equation}
where input image size is $H \times W$ with $C$ categories, and $P_{s}= (F\circ G)(x_s)$ is the softmax segmentation map produced by the classifier.\par

\textbf{Step 2: Generation of adversarial features. } 
The adversarial features $f_{s*/t*}$ are initialized with $f_{s/t}$ extracted by $G$ from $x_{s/t}$, and iteratively updated with our proposed I-FGSPM combining several attack objectives. These perturbed features are designed to confuse the discriminator and the classifier with our tailored attack objectives.\par

\textbf{Step 3: Training with adversarial features.} With adversarial features from step 2, it is crucial to set proper training objectives to defense against the perturbations and enable the classifier to produce consistent predictions. Besides, robust classifier and discriminator can contiguously generate confusing adversarial features for further training.\par

During training, we freeze $G$ after step 1, and alternate step 2 and step 3 to obtain a robust classifier against domain shift as well as category-conditional shift. We detail the step 2 and step 3 in the following sections.

\subsection{Generation of Adversarial Features}
In this part, we first introduce the attack objectives and then propose our {Iterative Fast Gradient Sign Preposed Method (I-FGSPM)} for combining multiple attack objectives. \par

\textbf{Attack objectives.} On the one hand, the generated perturbations are supposed to extrapolate the features towards domain-invariant regions. Therefore, they are expected to confuse the discriminator which aims to distinguish source domain from the target one by minimizing the loss function in Eq.~(\ref{adv loss}), 
so that the gradient of $ L_{adv}(P)$ is capable of producing perturbations that help fill in the domain gap.
\begin{equation}
    \begin{aligned}
        L_{adv}(P) = - \mathbb{E}[\log(D(P_s))] - \mathbb{E}[\log(1- D(P_t))].
    \end{aligned}
    \label{adv loss}
\end{equation}

On the other hand, to further improve the robustness of the classifier, the adversarial features should capture the vulnerability of the model (e.g., the tendency of classifier to collapse into head classes). In this regard, we conduct an adversarial attack on segmentation classifier and employ the Lov\'asz-Softmax~\cite{berman2018lovasz} as our attack objective in Eq~(\ref{seg loss}). Since the perturbations are actually hard examples for the classifier, they carry rich information of the failure mode of the segmentation classifier. Lov\'asz-Softmax is a smooth version of the jaccard index and we empirically show that our attack objective can produce proper segmentation perturbations as well as boosting the adaptability of the model.
\begin{equation}
    \label{seg loss}
    \begin{aligned}
        L_{seg}(P_s, y_s) = \text{Lov\'asz-Softmax}(P_s, y_s).
    \end{aligned}
\end{equation}
In addition, excessive perturbations might degenerate the semantic information of feature maps, so that we control the $L_2$-distance between the original features and their perturbed copies to self-adaptively constraint their divergence.
Eventually, we accumulate gradient maps from all attack objectives and generate adversarial features with our proposed Iterative Fast Gradient Sign preposed Method (I-FGSPM).\par
\textbf{Original I-FGSM.} 
While we can follow the practice in \cite{liu2019transferable} to directly regard the gradients as perturbations, we have empirically found that this strategy may suffer from gradient vanishing in our case. Instead, we draw a link from adversarial attack to generate more stable and reasonable perturbations.
Specifically, to generate the perturbations, we adopt the Iterative Fast Gradient Sign Method (I-FGSM)~\cite{kurakin2016adversarial} as Eq.~(\ref{I-FGSM}):
\begin{equation}
    \label{I-FGSM}
    \begin{aligned}
        f_{s*}^{k+1} =
         & f_{s*}^{k} + \epsilon \cdot \text{sign}(\beta_1 \nabla_{f_{s*}^k}
        L_{seg}(P_{s*}^{k}, y_s)                                             \\&- \beta_2 \nabla_{f_{s*}^k} L_2(f_{s*}^{k}, f_{s}) +
        \beta_3 \nabla_{f_{s*}^k} L_{adv}(P_{s*}^{k})),
    \end{aligned}
\end{equation}
where $\beta_1$, $\beta_2$ and $\beta_3$ indicate the hyper-parameters to balance the gradients values from different attack objectives and $\epsilon$ represents the magnitude of the overall perturbation. We repeat this generating process for $K$ iterations with $k\in \{0,1,\cdots,K-1\}$. It is noteworthy that $f_{s*}^0 = f_{s}$.

\begin{figure}
    \centering
    \includegraphics[width=0.8\linewidth]{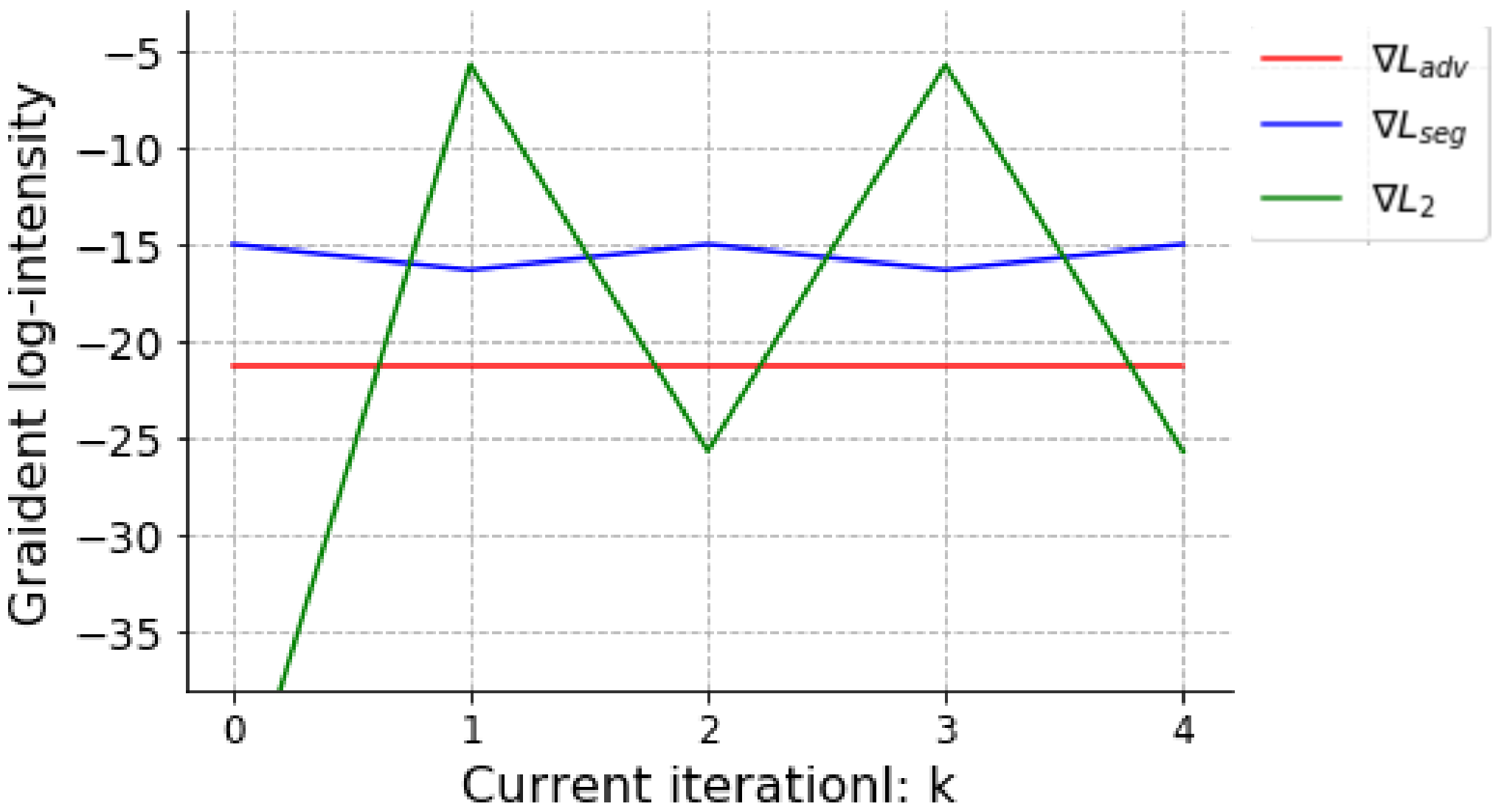}
    \caption{Gradient log-intensity tendencies with I-FGSM method in generation procedure.}
    \label{fig:loss_fluctuation}
\end{figure}

However, this practice also raises some concerns when we execute I-FGSM under the circumstance of multiple adversarial attack objectives. Such concerns are attributed to the significant gradient gaps among different attack objectives. It is worth mentioning that, at each iteration, the final signs of the accumulated gradients are indeed dominated by one of the attack objectives.
As illustrated in Figure~\ref{fig:loss_fluctuation}, we plot the gradient log-intensity of each attack objective by using Eq.~(\ref{I-FGSM}) to obtain adversarial features. In Figure~\ref{fig:loss_fluctuation}, the gradients of $L_{seg}$ and $L_2$ alternatively surpasses the others overwhelmingly with at least several orders of magnitude and therefore determine the final signs. Furthermore, the gradient value of a specific attack objective fluctuates by varying iterations and does not appear proportional tendency with its counterparts, so that it is not trivial to balance the gradient perturbations by simply adjusting the trade-off constants.

\textbf{Our I-FGSPM.} To this end, we propose the Iterative Fast Gradient Sign Preposed Method (\textbf{I-FGSPM}) to fully exploit the contributions of each individual attack objective. Rather than placing the sign operator at the end of the overall gradient fusion which suffers from the gradient domination issue, we instead put ahead the sign calculations of each adversarial gradient and then balance these signed perturbations with intensity $\epsilon$. The procedure is formulated as Eq.~(\ref{ML I-FGSM tar}) and (\ref{ML I-FGSM src}) for target and source perturbations respectively.
\begin{equation}
    \begin{aligned}
        f_{t*}^{k+1} =
        f_{t*}^{k} & + \epsilon_1 \text{sign}( \nabla_{f_{t*}^{k}}L_{adv}(P_{t*}^{k})) \\&-\epsilon_2 \text{sign}(\nabla_{f_{t*}^k}L_2(f_{t*}^{k}, f_{t})),
    \end{aligned}
    \label{ML I-FGSM tar}
\end{equation}
\begin{equation}
    \begin{aligned}
        f_{s*}^{k+1} =
        f_{s*}^{k} & + \epsilon_1 \text{sign}(\nabla_{f_{s*}^{k}}L_{adv}(P_{s*}^{k}))   \\
                   & -\epsilon_2  \text{sign}(\nabla_{f_{s*}^k}L_2(f_{s*}^{k}, f_{s})) \\&+ \epsilon_3  \text{sign}( \nabla_{f_{s*}^k}L_{seg}(P_{s*}^{k}, y_s)).
    \end{aligned}
    \label{ML I-FGSM src}
\end{equation}
\begin{table*}[t]
    \centering
    \caption{Results of adapting GTA5 to Cityscapes. The tail classes are highlighted in \textcolor{blue}{blue}. The top and bottom parts correspond to VGG-16 and ResNet-101 based model separately.
    }

    \label{table1}

    \begin{small}
        \setlength{\tabcolsep}{0.8mm}{
            \begin{tabularx}{\linewidth}{c|ccccccccccccccccccc|c}
                \hline \hline
                Method                              & \rotatebox{90}{road} & \rotatebox{90}{sidewalk\ } & \rotatebox{90}{building} & \rotatebox{90}{\textcolor{blue}{wall}} & \rotatebox{90}{\textcolor{blue}{fence}} & \rotatebox{90}{\textcolor{blue}{pole}} & \rotatebox{90}{\textcolor{blue}{light}} & \rotatebox{90}{\textcolor{blue}{sign}} & \rotatebox{90}{veg} & \rotatebox{90}{\textcolor{blue}{terrain}} & \rotatebox{90}{sky} & \rotatebox{90}{\textcolor{blue}{person}} & \rotatebox{90}{\textcolor{blue}{rider}} & \rotatebox{90}{car} & \rotatebox{90}{\textcolor{blue}{truck}} & \rotatebox{90}{\textcolor{blue}{bus}} & \rotatebox{90}{\textcolor{blue}{train}} & \rotatebox{90}{\textcolor{blue}{mbike}} & \rotatebox{90}{\textcolor{blue}{bike}} & mIoU          \\
                \hline
                ASN~\cite{tsai2018learning} & 87.3 & 29.8 & 78.6 & 21.1 & 18.2 & 22.5 & 21.5 & 11.0 & 79.7 & 29.6 & 71.3 & 46.8 & 6.5 & 80.1 & 23.0 & 26.9 & 0.0 & 10.6 & 0.3 & 35.0 \\
                CLAN~\cite{luo2019taking} & 88.0 & 30.6 & 79.2 & 23.4 & 20.5 & 26.1 & 23.0 & 14.8 & 81.6 & 34.5 & 72.0 & 45.8 & 7.9 & 80.5 & 26.6 & 29.9 & 0.0 & 10.7 & 0.0 & 36.6 \\
                \rowcolor{gray!30}Ours & 88.4 & 34.2 & 77.6 & 23.7 & 18.3 & 24.8 & 24.9 & 12.4 & 80.7 & 30.4 & 68.6 & 48.9 & 17.9 & 80.8 & 27.0 & 27.2 & 6.2 & 19.1 & 10.2 & 38.0\\
                \hline
                Source Only                         & 75.8                 & 16.8                       & 77.2                     & 12.5                 & 21.0                  & 25.5                 & 30.1                                    & 20.1                 & 81.3                & 24.6                    & 70.3                & 53.8                                     & 26.4                                    & 49.9                & 17.2                                    & 25.9                                  & 6.5                                     & 25.3                                    & 36.0                                   & 36.6          \\
                \small{ASN~\cite{tsai2018learning}} & 86.5                 & 25.9                       & 79.8                     & 22.1                 & 20.0                  & 23.6                 & 33.1                                    & 21.8                 & 81.8                & 25.9                    & 75.9                & 57.3                                     & 26.2                                    & 76.3                & 29.8                                    & 32.1                                  & 7.2                            & 29.5                                    & 32.5                                   & 41.4          \\
                \small{CLAN~\cite{luo2019taking}}   & 87.0                 & 27.1                       & 79.6                     & 27.3                 & 23.3                  & 28.3                 & 35.5                           & 24.2        & 83.6                & 27.4                    & 74.2                & 58.6                                     & 28.0                                    & 76.2                & 33.1                                    & 36.7                                  & 6.7                                     & 31.9                                    & 31.4                                   & 43.2          \\
                \small{AdvEnt\cite{vu2019advent}}   & \textbf{89.9}        &36.5              & \textbf{81.6}            & 29.2        & 25.2         & \textbf{28.5}        & 32.3                                    & 22.4                 & \textbf{83.9}       & 34.0                    & 77.1                & 57.4                                     & 27.9                                    & 83.7                & 29.4                                    & 39.1                                  & 1.5                                     & 28.4                                    & 23.3                                   & 43.8          \\
                \rowcolor{gray!13}ASN + Weighted CE                   & 82.8                 &  \textbf{42.4}                       & 77.1                     & 22.6                 & 21.8                  & 28.3                 & \textbf{35.9}                                    & \textbf{27.4}                 & 80.2                & 25.0                    & 77.2                & 58.1                                     & 26.3                                    & 59.4                & 25.7                                    & 32.7                                  & 3.6                                     & 29.0                                    & 31.4                                   & 41.4          \\
                \rowcolor{gray!13}ASN + Lov\'asz                      & 88.0                 & 28.6                       & 80.7                     & 23.6                 & 14.8                  & 25.9                 & 33.3                                    & 19.6                 & 82.8                & 31.1                    & 74.9                & 58.1                                     & 24.6                                    & 72.6                & 34.2                                    & 31.2                                  & 0.0                                     & 24.9                                    & 36.4                                   & 41.3          \\
                \rowcolor{gray!30}Ours & 85.6 & 32.8 & 79.0 & \textbf{29.5} & \textbf{25.5} & 26.8 & 34.6 & 19.9 & 83.7 & \textbf{40.6} & \textbf{77.9} & \textbf{59.2} & \textbf{28.3} & \textbf{84.6} & \textbf{34.6} & \textbf{49.2} & \textbf{8.0} & \textbf{32.6} & \textbf{39.6} & \textbf{45.9}\\
                \toprule[1pt]
            \end{tabularx}}
    \end{small}
    \vspace{-0.4cm}
\end{table*}
\subsection{Training with Adversarial Features}
Now, we are equipped with adversarial features which can reduce the domain gap and capture the vulnerability of the classifier. 
To obtain a domain-invariant classifier $F$ and a robust domain discriminator $D$, we should design proper constraints that can guide the learning process to utilize these adversarial features to train $F$ and $D$.\par

For this purpose, the solution appears straightforward for the source domain since we still hold the strong supervision $y_s$ for its adversarial features $f_{s*}$. On the contrary, when it comes to the unlabeled target domain, we are supposed to explore other supervision signals to satisfy the goal. Our considerations are two folds. First, we follow the practice in \cite{liu2019transferable} that forces the classifier to make consistent predictions for $f_{t}$ and $f_{t*}$ as follow:
\begin{equation}
    \label{cst loss}
    L_{cst}(P_{t}, P_{t*}) =  \mathbb{E}[\left \| P_t - P_{t*} \right \|_2].
\end{equation}
Noted that this action does not guarantee the discriminative and reductive information for specific tasks. 
Instead, as the perturbations intend to confuse the classifier, the prediction maps of adversarial features are empirically subject to have more uncertainty with increasing entropy.
To address this issue, we draw on the idea of the entropy minimization technique~\cite{springenberg2015unsupervised,long2018transferable} as Eq~(\ref{ent loss}) to provide extra supervision, which can be viewed as a soft-assignment variant of the pseudo-label cross entropy loss~\cite{vu2019advent}. 
\begin{equation}
    \label{ent loss}
    \begin{aligned}
        L_{ent}(P) =  \mathbb{E}[ \frac{-1}{\log(C)}\sum_{c=1}^C P^{(h,w,c)}\log P^{(h,w,c)}].
    \end{aligned}
\end{equation}
Finally, by combining the objectives in (\ref{seg loss}), (\ref{cst loss}) and (\ref{ent loss}), we are capable of obtaining robust and discriminative classifier $F$ as follow, where $\alpha_1$, $\alpha_2$ and $\alpha_3$ are trade-off factors:
\begin{equation}
    \centering
    \small
    \begin{aligned}
        \min_F L_{cls} & = L_{seg}(P_{s*}, y_s) + L_{seg}(P_{s}, y_s) 
        + \alpha_1L_{cst}(P_t, P_{t*}) \\&+ \alpha_2L_{ent}(P_t) + \alpha_3L_{ent}(P_{t*}).
    \end{aligned}
\end{equation}\par
In addition, we conduct a similar procedure to defense against domain-related perturbations, which forces the discriminator $D$ to assign the same domain labels for the mutated features with respect to their original ones. Furthermore, it is beneficial for the discriminator to contiguously generate perturbations that extrapolate the features towards more domain-invariant regions and then bridge the domain discrepancy more effectively.

\section{Experiments}
\subsection{Dataset}
We evaluate our method along with several state-of-the-art algorithms on two challenging \textit{synthesized-2-real} UDA benchmarks, i.e., \textbf{GTA5 $\rightarrow$ Cityscapes} and \textbf{SYNTHIA $\rightarrow$ Cityscapes}. Cityscapes is a real-world image dataset, consisting of 2,975 images for training and 500 images for validation. GTA5 contains 24,966 synthesized frames captured from  the video game. We use the 19 classes of GTA5 in common with the Cityscapes for adaptation. SYNTHIA is a synthetic urban scenes dataset with 9,400 images. Similar to \citeauthor{vu2019advent}~\shortcite{vu2019advent}, We train our model with 16 common classes in both SYNTHIA and Cityscapes, and evaluate the performance on 13-class subsets.

\begin{table*}[t]
    \centering
    \caption{Results of adapting SYNTHIA to Cityscapes. The tail classes are highlighted in \textcolor{blue}{blue}.
    }
    \label{table2}
    \begin{small}
        \setlength{\tabcolsep}{2.1mm}{
            \begin{tabularx}{\linewidth}{c|ccccccccccccc|c}
                \hline \hline
                Method                      & \rotatebox{90}{road} & \rotatebox{90}{sidewalk\ } & \rotatebox{90}{building} & \rotatebox{90}{\textcolor{blue}{light}} & \rotatebox{90}{\textcolor{blue}{sign}} & \rotatebox{90}{veg} & \rotatebox{90}{sky} & \rotatebox{90}{person} & \rotatebox{90}{\textcolor{blue}{rider}} & \rotatebox{90}{car} & \rotatebox{90}{\textcolor{blue}{bus}} & \rotatebox{90}{\textcolor{blue}{mbike}} & \rotatebox{90}{\textcolor{blue}{bike}} & $\text{mIoU}_{13}$ \\
                \hline
                ASN~\cite{tsai2018learning} & 78.9 & 29.2 & 75.5 & 0.1 & 4.8 & 72.6 & 76.7 & 43.4 & 8.8 & 71.1 & 16.0 & 3.6 & 8.4 & 37.6\\
                CLAN~\cite{luo2019taking} & 80.4 & 30.7 & 74.7 & 1.4 & 8.0 & 77.1 & 79.0 & 46.5 & 8.9 & 73.8 & 18.2 & 2.2 & 9.9 & 39.3 \\
                \rowcolor{gray!30}Ours & 82.9 & 31.4 & 72.1 & 10.4 & 9.7 & 75.0 & 76.3 & 48.5 & 15.5 & 70.3 & 11.3 & 1.2 & 29.4 & 41.1 \\
                \hline
                Source Only                 & 55.6                 & 23.8                       & 74.6                     & 6.1                                     & 12.1                                   & 74.8                & 79.0                & 55.3                   & 19.1                                    & 39.6                & 23.3                                  & 13.7                                    & 25.0                                   & 38.6               \\
                ASN~\cite{tsai2018learning} & 79.2                 & 37.2                       & 78.8                     & 9.9                                     & 10.5                                   & 78.2                & 80.5                & 53.5                   & 19.6                                    & 67.0                & 29.5                                  & 21.6                                    & 31.3                                   & 45.9               \\
                CLAN~\cite{luo2019taking}   & 81.3                 & 37.0                       & \textbf{80.1}            & 16.1                                    & 13.7                                   & 78.2                & 81.5                & 53.4                   & 21.2                                    & 73.0                & 32.9                                  & 22.6                                    & 30.7                                   & 47.8               \\
                AdvEnt~\cite{vu2019advent}  & \textbf{87.0}        & \textbf{44.1}              & 79.7                     & 4.8                                     & 7.2                                    & 80.1                & \textbf{83.6}       & 56.4                   & \textbf{23.7}                           & 72.7                & 32.6                                  & 12.8                                    & 33.7                                   & 47.6               \\
                \rowcolor{gray!13}ASN + Weighted CE           & 74.9                 & 37.6 & 78.1                     & 10.5                                    & 10.2                                   & 76.8                & 78.3                & 35.3                   & 20.1                                    & 63.2                & 31.2                                  & 19.5                                    & 43.3                                   & 44.5               \\
                \rowcolor{gray!13}ASN + Lov\'asz              & 77.3                 & 40.0                       & 78.3                     & 14.4                                    & 13.7                                   & 74.7                & 83.5                & 55.7                   & 20.9                                    & 70.2                & 23.6                                  & 19.3                                    & 40.5                                   & 47.1               \\
                \rowcolor{gray!30}Ours                        & 86.4                 & 41.3                       & 79.3                     & \textbf{22.6}                           & \textbf{17.3}                          & \textbf{80.3}       & 81.6                & \textbf{56.9}          & 21.0                                    & \textbf{84.1}       & \textbf{49.1}                         & \textbf{24.6}                           & \textbf{45.7}                          & \textbf{53.1}      \\
                \toprule[1pt]
            \end{tabularx}}
    \end{small}
    \vspace{-0.4cm}
\end{table*}
\subsection{Implementations details} We use PyTorch for implementation. Similar to~\citeauthor{tsai2018learning}~\shortcite{tsai2018learning}, we utilize the DeepLab-v2 \cite{chen2017deeplab} as our backbone segmentation network. We employ Atrous Spatial Pyramid Pooling (ASPP) as classifier followed by an up-sampling layer with softmax output. For domain discriminator $D$, we use the one in DCGAN~\cite{radford2015unsupervised} but exclude batch normalization layers. Our experiments are based on two different network architectures: VGG-16~\cite{simonyan2014very} and ResNet-101~\cite{he2016deep}.
During training, we use SGD~\cite{bottou2010large} for $G$ and $C$ with momentum 0.9, learning rate $2.5 \times 10^{-4}$ and weight decay $10^{-4}$. We use Adam~\cite{kingma2014adam} with learning rate $10^{-4}$ to optimize $D$. And we follow the polynomial annealing procedure~\cite{chen2017deeplab} to schedule the learning rate. When generating adversarial features, the iteration $K$ of I-FGSPM is set to 3. Note that we set the $\epsilon_1$, $\epsilon_2$ and $\epsilon_3$ in Eq.~(\ref{ML I-FGSM tar}) and (\ref{ML I-FGSM src}) as $0.01$, $0.002$ and $0.011$ separately. $\alpha_1$, $\alpha_2$ and $\alpha_3$ are $0.2$, $0.002$ and $0.0005$ separately.

\begin{figure}[t]
    \centering
    \setlength{\abovecaptionskip}{0.05cm}
    \setlength{\belowcaptionskip}{-0.3cm}
    \includegraphics[width=\linewidth]{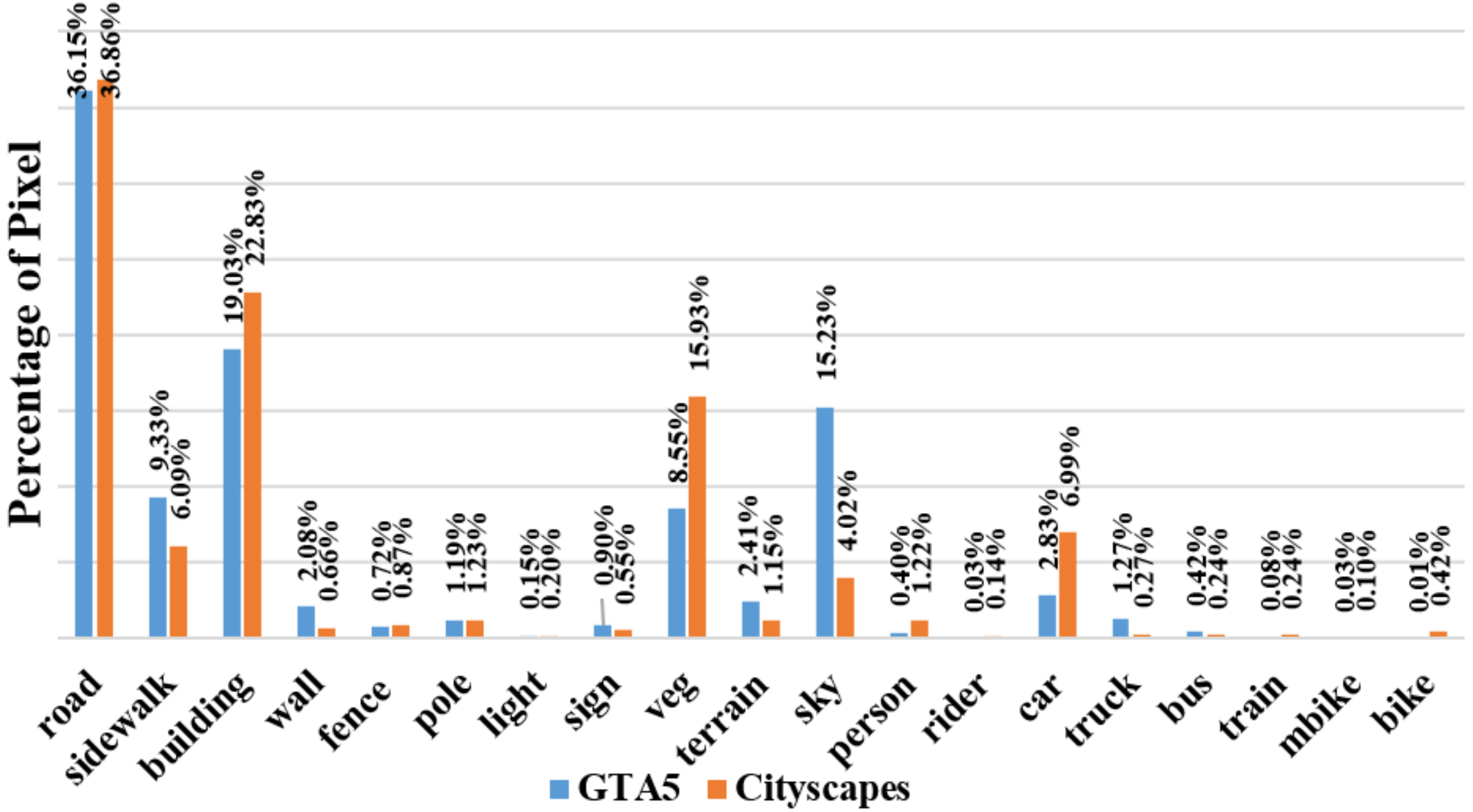}
    \caption{Category distribution on GTA5 $\rightarrow$ Cityscapes.}
    \label{fig:distribution-gta5}
\end{figure}

\subsection{Result Analysis}
We compare our model with several state-of-the-art domain adaptation methods on semantic segmentation performance in terms of mIoU. 
Table~\ref{table1} shows that our ResNet-101 based model brings +9.3\% gain compared to source only model on GTA5 $\rightarrow$ Cityscapes. Besides, our method also outperforms state-of-the-arts over +1.4\% and +2.1\% in mIoU on VGG-16 and ResNet-101 separately. To further illustrate the {effectiveness} of our method on tail classes,  we show the marginal category distributions counted in 19 common classes on GTA5 and Cityscapes datasets in Figure~\ref{fig:distribution-gta5}, and highlight the tail classes with \textcolor{blue}{blue} in Table~\ref{table1}.
{For example, the category ``bike" accounts for only 0.01\% ratio in the GTA5 category distribution, and the ResNet-101 based adversarial alignment methods suffer from  a huge performance degradation compared to the source only model. Specifically, AdvEnt can deliver a +7.2\% performance improvement on average, but the category ``bike'' itself suffers 12.7\% performance degradation. On the contrary, our approach can still improve the performance of the ``bike'' category by benefiting from the pointwise perturbation strategy. In fact, our framework can achieve the best performance at the majority of tail categories, showing the effectiveness of our algorithm in mitigating the category-conditional shift.}\par

\begin{figure*}[t]
    \centering
    \includegraphics[width=\linewidth]{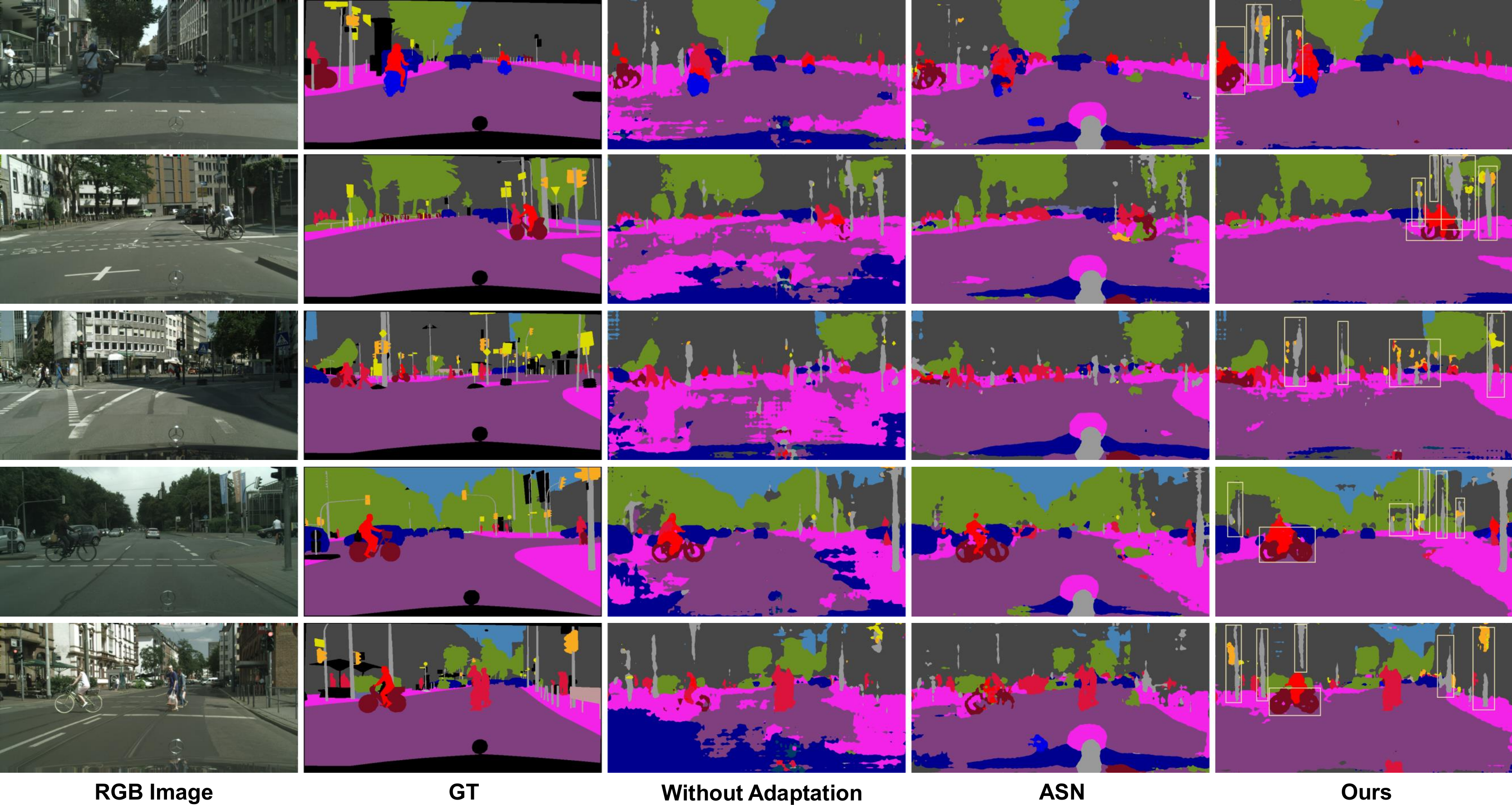}
    \caption{Qualitative results of UDA segmentation for SYNTHIA $\rightarrow$ Cityscapes. Along with each target image and its corresponding ground truth, we present the results of source only model (without adaptation), ASN and ours respectively.}
    \label{fig:Qualitative_result}
\end{figure*}

Table~\ref{table2} provides the comparative performance on SYNTHIA $\rightarrow$ Cityscapes. SYNTHIA has significantly different layouts as well as viewpoints compared {to} Cityscapes, and less training samples than GTA5. Hence, models trained in SYNTHIA might suffer from serious domain shift when generalized into Cityscapes. 
It is noteworthy that our adversarial perturbation framework generates hard examples that strongly resist adaptation, thus our model can efficiently improve performance in the difficult task by considering these augmented features.
As a result, our method significantly outperforms the state-of-the-art methods by +1.8\% and +5.5\% in mIoU based on VGG-16 and ResNet-101 separately. Specifically, even when compared to CLAN method, which aims at aligning category-level joint distribution, our framework {still} achieves higher performance on tail classes. Some qualitative results are presented in Figure~\ref{fig:Qualitative_result}.\par

\begin{figure}[t]
    \centering
    \setlength{\abovecaptionskip}{0.05cm}
    \setlength{\belowcaptionskip}{-0.2cm}
    \includegraphics[width=1\linewidth]{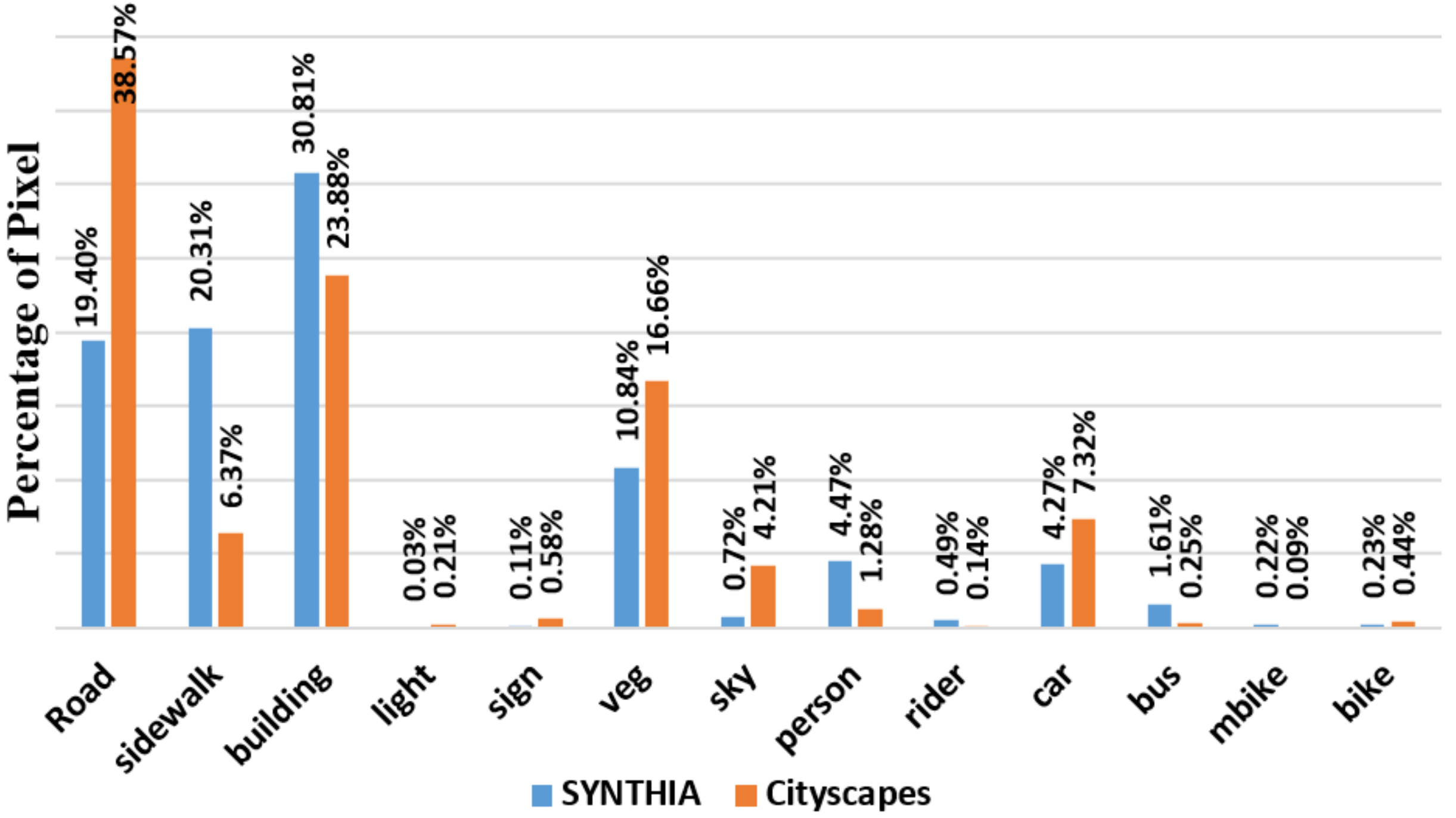}
    \caption{Category distribution on SYNTHIA $\rightarrow$ Cityscapes.}
    \label{fig:distribution-syn}
\end{figure}

Furthermore, we re-implement ASN with some category balancing mechanisms (e.g., weighted cross entropy and Lov\'asz-Softmax loss) based on ResNet-101 for fair comparison. As shown in Table~\ref{table1} and \ref{table2}, we show that only ASN + Lov\'asz brings +1.2\% gain in SYNTHIA $\rightarrow$ Cityscapes, while others even suffer from performance degradation. As shown in Figure~\ref{fig:distribution-gta5} and \ref{fig:distribution-syn}, marginal category distributions are varying across domains, and thus re-weighting mechanisms can not guarantee adaptability on the target domain. 

\section{Ablation Study}
\textbf{Different Attack Methods.} A basic problem of our framework is how to generate proper perturbations. We compare several attack methods widely used in adversarial attack community and their modified sign-preposed versions. Specifically, we compare our proposed I-FGSPM with I-FGSM and modified sign-preposed version of FGSM~\cite{goodfellow2014explaining} as well as Momentum I-FGSM (MI-FGSM)~\cite{dong2018boosting}. Furthermore, we also provide a ``None" version without any attacks.
As illustrated in Table~\ref{tab:attack_method}, ResNet-101 based adversarial attack methods bring obvious gain against ``None'' version. With sign-preposed operation, our I-FGSPM achieves +1.3\% improvement compared to I-FGSM. FGSPM is the non-iterative version of our I-FGSPM and achieves comparable performance against I-FGSPM. Note that though MI-FGSM achieves remarkable results in adversarial attacks area, its sign-preposed version MI-FGSPM might excessively enlarge the divergence between original features with adversarial features, and causes performance degradation when employed by our framework.
\begin{table}
    \centering
    \setlength{\abovecaptionskip}{-0.1cm}
    \caption{Evaluation on different attack methods.}
    \begin{small}
        \begin{tabular}{c|c}
            \bottomrule[1pt]
            Attack Method & $\text{mIoU}_{13}$ (SYNTHIA) \\
            \hline
            None          & 44.8                        \\
            I-FGSM        & 51.8                        \\
            FGSPM         & 52.9                        \\
            MI-FGSPM      & 52.2                        \\
            \rowcolor{gray!30}I-FGSPM (Ours) & \textbf{53.1}               \\
            \toprule[0.8pt]
        \end{tabular}
    \end{small}
    \label{tab:attack_method}
    \vspace{-0.1cm}
\end{table}\par
\textbf{Different perturbing layers.} One natural question is whether it is better to perturb the input or the hidden layers of model. \citeauthor{szegedy2013intriguing}~\shortcite{szegedy2013intriguing} reported that adversarial perturbations yield the best regularization when applied to the hidden layers. Our experiments with ResNet-101 shown in Table~\ref{tab:different_layer} also verify that perturbing in feature-level achieves the best result. These might boil down to that the activation of hidden units can be unbounded and very large when perturbing the hidden layers \cite{goodfellow2014explaining}. We also find that perturbing deeper hidden layers can further benefit our framework.
\begin{table}
    \centering
    \setlength{\abovecaptionskip}{-0.1cm}
    \caption{Evaluation on different perturbing layers.}
    \begin{small}
        \begin{tabular}{c|c}
            \bottomrule[1pt]
            Layer               & $\text{mIoU}_{13}$ (SYNTHIA) \\
            \hline
            Pixel-level         & 50.4                        \\
            \rowcolor{gray!10}After layer1        & 45.0                        \\
            \rowcolor{gray!10}After layer2        & 49.8                        \\
            \rowcolor{gray!10}After layer3        & 50.6                        \\
            \rowcolor{gray!30}After layer4 (Ours) & \textbf{53.1}               \\
            \toprule[0.8pt]
        \end{tabular}
    \end{small}
    \label{tab:different_layer}
    \vspace{-0.45cm}
\end{table}
\begin{table}
    \centering
    \setlength{\abovecaptionskip}{-0.1cm}
    \caption{Ablation studies of each component. ``S'' represents our strategy as discussed in step 1 while ``ASN'' indicates that our network weights are pre-trained by ASN in step 1.}
    \setlength{\tabcolsep}{0.4mm}{
        \begin{small}
            \begin{tabular}{c|ccc|c|c}
                \bottomrule[1pt]
                Base & Perturbation & Lov\'asz & Entropy & mIoU (GTA5)    & $\text{mIoU}_{13}$ (SYN) \\
                \hline
                S& & &  & 36.6& 38.6\\
                S& & $\surd$  &  & 35.0& 41.3\\
                S& & & $\surd$ & 41.8 & 42.5\\
                S& & $\surd$  & $\surd$ & 38.5 & 44.8 \\
                \rowcolor{gray!13}S& $\surd$  & & & 41.7  & 45.7   \\
                \rowcolor{gray!13}S& $\surd$ & $\surd$  & & 44.6  & 49.9  \\
                \rowcolor{gray!13}S& $\surd$ &  & $\surd$ & 43.6 & 47.0  \\
                \rowcolor{gray!30}S& $\surd$ & $\surd$ & $\surd$ & \textbf{45.9} & \textbf{53.1} \\
                \hline
                ASN & & & & 41.4 & 45.9 \\
                ASN & & $\surd$ & $\surd$ & 42.3 & 47.4 \\
                \rowcolor{gray!30}ASN & $\surd$ & $\surd$ & $\surd$ & 45.2 & 52.9 \\ 
                \toprule[0.8pt]
            \end{tabular}
        \end{small}}
    \label{tab:ablation_study}
    \vspace{-0.5cm}
\end{table}\par
\textbf{Component Analysis.} We study how each component affects overall performance in terms of mIoU based on ResNet-101. As shown in the top part of Table~\ref{tab:ablation_study}, starting with source only model trained with Lov\'asz-Softmax, we notice that the effect of Lov\'asz-Softmax loss varies across different UDA tasks, which might depend on how different the marginal distributions across two domains are. Entropy minimization strategy can  bring  improvement on both benchmarks but lead to strong class biases, which has been verified in AdvEnt \cite{vu2019advent}, while our overall model not only significantly lifts mIoU, but also remarkably alleviates category biases specially for tail classes.\par
As illustrated in the bottom part of Table~\ref{tab:ablation_study}, we consider our basic training strategy in step 1 as a component, and replace it with ASN.
By cooperating with our perturbations strategy, ours + ASN brings +3.8\% and +7.0\% gain, while ASN + Lov\`asz + Entropy only gets +0.9\% and +1.5\% improvement against ASN on GTA5 to Cityscapes and SYNTHIA to Cityscapes separately.
A possible reason is that ASN can shape the feature extractor biased towards the head classes and miss representations from tail classes.\par

\section{Conclusion}
In this paper, we reveal that adversarial alignment based segmentation DA might be dominated by head classes and fail to capture the adaptability of different categories evenly. To address this issue, we proposed a novel framework that iteratively exploits our improved I-FGSPM  to extrapolate the perturbed features towards more domain-invariant regions and defenses against them via an adversarial training procedure. The virtues of our method lie in not only the adaptability of model but that it circumvents the intervention among different categories. Extensive experiments have verified that our approach significantly outperforms the state-of-the-arts, especially for the hard tail classes. 

{
    \small
\bibliographystyle{aaai}
\bibliography{egbib2}
}
\end{document}